\documentclass[journal]{IEEEtran}

\usepackage{xcolor,soul,framed} 

\colorlet{shadecolor}{yellow}
\usepackage[pdftex]{graphicx}
\graphicspath{{../pdf/}{../jpeg/}}
\DeclareGraphicsExtensions{.pdf,.jpeg,.png}

\usepackage[cmex10]{amsmath}
\usepackage{array}
\usepackage{mdwmath}
\usepackage{mdwtab}
\usepackage{eqparbox}
\usepackage{url}

\usepackage{amssymb}
\usepackage{multirow}
\newcommand{\RNum}[1]{\uppercase\expandafter{\romannumeral #1\relax}}

\begin{document}
\bstctlcite{IEEEexample:BSTcontrol}
    \title{Temporal Superimposed Crossover Module for Effective Continuous Sign Language}
  \author{
  
      

    \IEEEauthorblockN{Qidan~Zhu$^{1}$, 
    Jing~Li$^{1*}$, 
    Fei~Yuan$^2$, 
    Quan~Gan$^1$}
    
    \IEEEauthorblockA{$^1$ College of Intelligent Systems Science and Engineering, Harbin Engineering University, Harbin, 150001, China}
    \IEEEauthorblockA{$^2$ Northwest Institute of Mechanical and Electrical Engineering, Xianyang, 712099, China}
 
 \thanks{*Corresponding author\par
Email addresses: \par
zhuqidan@hrbeu.edu.cn (Qidan Zhu), \par
ljing@hrbeu.edu.cn (Jing Li), \par
bohelion@hrbeu.edu.cn (Fei Yuan), \par
gquan@hrbeu.edu.cn (Quan Gan)}
  }

\maketitle

\begin{abstract}
The ultimate goal of continuous sign language recognition(CSLR) is to facilitate the communication between special people and normal people, which requires a certain degree of real-time and deploy-ability of the model. However, in the previous research on CSLR, little attention has been paid to the real-time and deploy-ability. In order to improve the real-time and deploy-ability of the model, this paper proposes a zero parameter, zero computation temporal superposition crossover module(TSCM), and combines it with 2D convolution to form a "TSCM+2D convolution" hybrid convolution, which enables 2D convolution to have strong spatial-temporal modelling capability with zero parameter increase and lower deployment cost compared with other spatial-temporal convolutions. The overall CSLR model based on TSCM is built on the improved ResBlockT network in this paper. The hybrid convolution of "TSCM+2D convolution" is applied to the ResBlock of the ResNet network to form the new ResBlockT, and random gradient stop and multi-level CTC loss are introduced to train the model, which reduces the final recognition WER while reducing the training memory usage, and extends the ResNet network from image classification task to video recognition task. In addition, this study is the first in CSLR to use only 2D convolution extraction of sign language video temporal-spatial features for end-to-end learning for recognition. Experiments on two large-scale continuous sign language datasets demonstrate the effectiveness of the proposed method and achieve highly competitive results.
\end{abstract}


%
\IEEEpeerreviewmaketitle


\section{Introduction}

\IEEEPARstart{T}{he} main purpose of sign language recognition is to translate sign language movements into natural languages that can be recognized by normal people, so as to reduce communication barriers between special people and normal people, and it can also be used as an application tool for human-computer interaction\cite{wei2020semantic}\cite{adaloglou2021comprehensive}\cite{du2022full}. In the last decade, video-based sign language recognition (VSLR) has developed significantly and become the mainstream of sign language recognition at present. According to the results of VSLR, it can be divided into: isolated word sign language recognition and CSLR, while in real scenarios, CSLR has greater application value\cite{cui2017recurrent}\cite{zhu2022multi}.\par

In the early stage, in CSLR, researchers used the combination of manual features and HMM to recognize sign language videos\cite{yang2016continuous}\cite{zhang2016chinese}. With the advent and development of CNN, CNN has replaced manual features with its excellent feature extraction capability and combined with HMM to form a hybrid model of "CNN+HMM"\cite{koller2017re}\cite{koller2016deep}. Later, due to the excellent performance of LSTM in temporal processing, combined with the feature extraction capability of CNN, it was applied to CSLR\cite{al2021deep}\cite{huang2021boundary}. In recent years, multimodal and multi-thread models have emerged in order to achieve higher recognition accuracy and to obtain richer feature information. Two-stream networks with one stream for RBG images and one stream for optical flow images have achieved good results in CSLR\cite{cui2019deep}. And in order to obtain more cues, hand shape, facial expressions, human posture, etc. can be used as the input of multimodal model\cite{zhou2020spatial}. However, these studies are aimed at improving the accuracy of CSLR models, and less research has been done on real-time and deploy-ability\cite{cheng2020fully}. The initial purpose of the sign language recognition model is to reduce communication barriers between special people and normal people. As a communication tool, there should be high requirements for the accuracy, real-time and deploy-ability of the model\cite{rastgoo2021sign}. VSLR is a challenging task to improve the real-time and deploy-ability of CSLR models because the input is video data and the spatial-temporal relationships between video frames need to be fully considered, resulting in a very computationally intensive recognition model.\par

\begin{figure*}
  \begin{center}
  \includegraphics[width=6.5in]{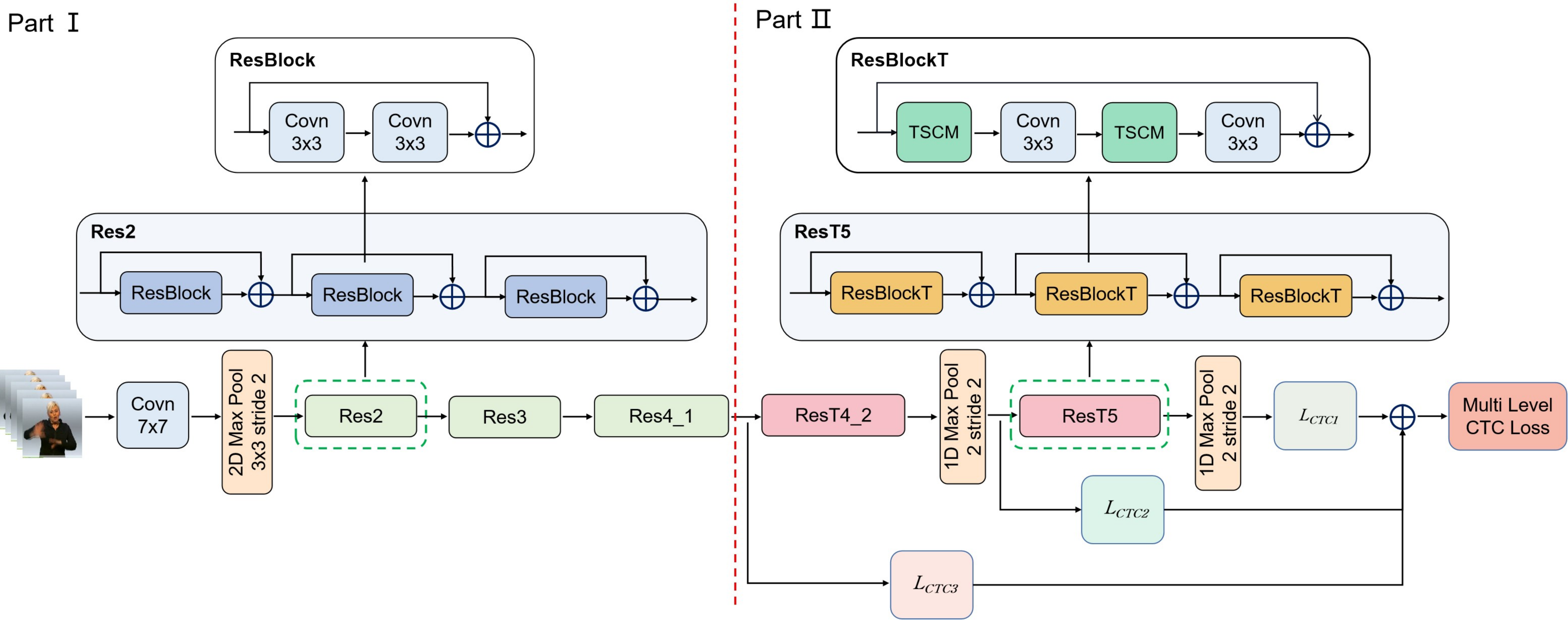}\\
  \caption{The framework of the ResNetT34 model based on the improved ResNet34, where Part1 is the original ResNet34 part, Part2 is the improved part, 1D-MaxPool is a down-sampling of the time dimension, and a random gradient stop is used for Part1 to reduce memory usage and training time during the training process.}\label{fig:ljxy1}
  \end{center}
\end{figure*}

To address the above problems, a zero-computation, zero-parameter TSCM is proposed to combine with 2D convolution to form a "TSCM+2D convolution" hybrid convolution, which can be flexibly inserted into existing classical classification models, and has strong universality, and can effectively model the spatial-temporal information of any video. In this paper, ResNet\cite{he2016deep} is improved by replacing the normal convolution of the residual branch in ResBlock with the proposed hybrid convolution, and the resulting new Block is named ResBlockT, and random gradient stop\cite{niu2020stochastic} and multi-level CTC loss\cite{zhu2022multi} are introduced for training, which reduces the word error rate (WER) of the final recognition while reducing the training memory usage. The proposed method has been tested on two publicly available continuous sign language datasets, and has achieved highly competitive results. \par

The main contributions of this paper are as follows:\par

\begin{itemize}
\item[$\bullet$] This paper proposes a zero-computation, zero-parameter TSCM that combines adjacent temporal data into one data, which can be flexibly and conveniently inserted into existing classical classification models to extend image recognition tasks to video recognition tasks.
\item[$\bullet$] This paper proposes a hybrid convolution formed by combining TSCM and 2D convolution, which can be flexibly and conveniently inserted into existing classical classification models, enabling 2D-CNNs to also handle 3D video data well.
\item[$\bullet$] In this paper, we improve the ResNet network, replace the ordinary convolution in the residual branch of ResBlcok with the hybrid convolution we proposed, form a new block called ResBlockT, and introduce random gradient stop and multi-level CTC loss to train the model, which reduces the use of training memory usage while reducing the final identified WER.
\item[$\bullet$] The improved network in this paper is the first end-to-end network in CSLR that uses only 2D convolution for spatial-temporal feature extraction. Experiments have been carried out on two publicly available continuous sign language datasets, and highly competitive results have been achieved.
\end{itemize}

\section{Related work}

\subsection {Continuous sign language recognition}
VSLR is the translation of continuous sign language videos into comprehensible written phrases or spoken words. While traditional methods usually extract manual features and then combine them with HMM\cite{yang2016continuous}\cite{talukdar2022vision} or dynamic time warping (DTW)\cite{zhang2014threshold} methods, the rise of CNNs has replaced the extraction of manual features. Koller et al.\cite{koller2018deep} embedded CNN end-to-end into HMM, and explained the output of CNN in the Bayesian framework. The CNN-HMM hybrid model combines the strong discrimination ability of CNN and the sequence modeling ability of HMMs. In RNN, both the high temporal modelling capability of LSTMs and the alignment approach of CTC\cite{graves2006connectionist} for non-aligned sequences are well suited for CSLR. Gao et al.\cite{gao2021rnn} proposed an effective RNN converter-based Chinese sign language processing method and designed a multi-level visual hierarchical transcription network with frame-level, lexical-level and phrase-level BiLstm to explore multi-scale visual semantic features. Min et al.\cite{min2021visual} proposed visual alignment constraints to enable end-to-end training of CSLR networks by enforcing feature extractors for more alignment supervision for prediction, which was used to address the overfitting problem of CTC in CSLR.\par

\begin{figure*}
  \begin{center}
  \includegraphics[width=6.5in]{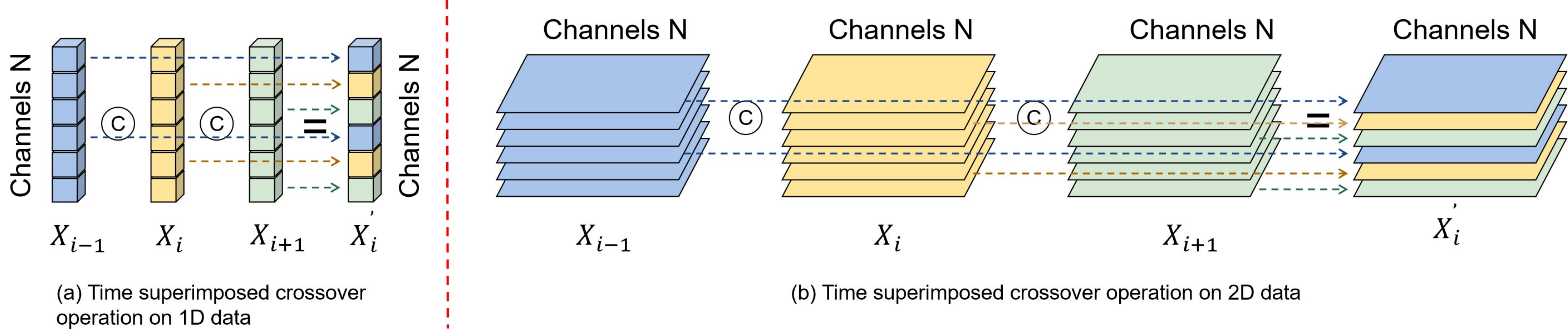}\\
  \caption{Figure (a)(b) shows the time superimposed crossover operation on 1D data and 2D data respectively, where $\copyright $ is the symbol of the channel crossover operation and $X_{t}^{'}$ is the result after the time superimposed crossover.}\label{fig:ljxy2}
  \end{center}
\end{figure*}

3D-CNN is one of the common methods to process video sequences and can be very effective in establishing spatial-temporal dependency, and have also been widely used in CSLR. Ariesta et al.\cite{ariesta2018sentence} proposed a sentence-level sign language method combining 3D-CNN and bidirectional recurrent neural network(Bi-RNN) for deep learning. 3D-CNN network has the disadvantages of being computationally intensive and bulky models, and in order to design a lightweight 3D-CNN network, the 3D convolution was decomposed into “2+1D” convolution. Han et al.\cite{han2022sign} used "2+1D-CNN" for feature extraction and proposed a lightweight spatial-temporal channel attention module. \par

The input of the CSLR model is no longer limited to the original sign language video data, and more diversified multimodal input is used to obtain more abundant feature information. Cui et al.\cite{cui2019deep} developed a CSLR system with recursive CNN on the multi-modal data of RGB frames and optical flow images. Zhou et al.\cite{zhou2020spatial} proposed a new CSLR multi clue framework and designed a spatial multi clue module with a self-contained attitude estimation branch to decompose spatial multi clue features.\par

In this paper, in order to improve the real-time and deploy-ability of the CSLR model, the main differences between our work and the above work are: 1) using only RGB images as input, improving on the classical classification model; 2) instead of establishing temporal relationship via LSTM or 3D-CNN, end-to-end training is performed by the proposed method using only 2D convolution for temporal feature extraction.\par

\subsection {Time module plugin}
The time module plugin is to insert the module into the existing 2D-CNN network, which can expand the 2D-CNN network from image classification task to video classification task, so that under the premise of maintaining the better real-time and deploy-ability of the 2D-CNN network, the spatial-temporal relationship can be efficiently modeled to achieve the balance of performance and accuracy. Wang et al.\cite{wang2018non} proposed a new neural network, which captures long-term dependencies through nonlocal operations. Lin et al.\cite{lin2019tsm} proposed a time shift module for hardware efficient video recognition, which moves part of the channel along the time dimension to exchange information with adjacent frames. Yang et al.\cite{yang2022stsm} proposed a spatial-temporal displacement module for efficient video recognition. This module moves some channels in the time dimension and space dimension of different channels, enabling the network to learn its spatial-temporal characteristics. Liu et al.\cite{liu2021tam} proposed a general time module, called Time Adaptive Module (TAM), to capture complex motion patterns in video, and proposed a powerful video architecture (TANet) based on the new time module. Su et al.\cite{nguyen2022vidconv} adopted a modern two-dimensional convolutional network for video action recognition and designed a new action recognition backbone. In this study, the time overlapping cross module method proposed by us is the most similar to the time module plugin, which establishes a more effective spatial-temporal relationship by overlapping the channel information of adjacent time series data.\par

\section{Methodology}
In this section, the technical details of the new TSCM, a plug-and-play module with zero computation and zero parameters, are first presented, which can be combined with 2D convolution to form a "TSCM+2D convolution" hybrid convolution that can effectively encode spatial-temporal features when embedded in the target network. It is then described how to embed the hybrid convolution into the existing 2D-CNN classical architecture, so that the 2D-CNN can perform the video recognition task well. In this paper, ResNet34 is used as an example for improvement, and the improved network is called ResNetT34, as shown in Figure 1.\par

\subsection {Temporal superposition crossover module}

The temporal superposition crossover module consists of two parts: temporal superposition and channel crossover, as shown in Figure 2. For each channel of adjacent temporal data, the proposed partial channel overlay crossover method is used to obtain temporal features in order to reduce the number of model parameters while ensuring the validity of the features. The details of temporal superposition and channel crossover are presented next, respectively.\par

The calculation process of traditional 1D convolution is as follows:\par

Let there exist a 1D time-series data $X$ of infinite length, where any three adjacent data can be denoted as $X_{i-1}$, $X_i$, $X_{i+1}$, $i\in[1,length(X)-2]$. The data is computed using a 1D convolution with a convolution kernel size of 3 and a weight of $W=(w_1,w_2,w_3)$. The convolution process is as follows:\par

\begin{equation}
Y_i = w_1X_{i-1}+w_1X_{i}+w_1X_{i+1}
\end{equation}

where $Y_i$ denotes the output of the convolution calculation. 1D The convolution calculation process is a summation process in which adjacent temporal data are multiplied by the corresponding weights according to the size of the convolution kernel.\par

\begin{figure*}
  \begin{center}
  \includegraphics[width=6.5in]{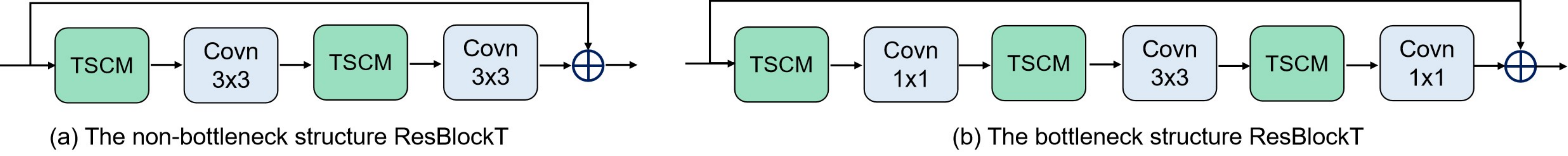}\\
  \caption{Figures (a) and (b) show the non-bottleneck structure and the bottleneck structure ResBlockT, respectively.}\label{fig:ljxy3}
  \end{center}
\end{figure*}

A straightforward approach to the temporal superposition problem is to superimpose adjacent temporal data in the channel dimension. That is, we can use a 1D convolution with a convolution kernel of size 1, an input channel number that is expanded by a factor of $n$ ($n$ is the number of superimposed data), and an output channel number that remains the same as the channel number of the original data $X_i$ to perform a convolution operation to achieve the same temporal modelling effect as in equation (1). Taking any three adjacent data as an example, let the superimposed data be $X_i^{'}=[X_{i-1},X_i,X_{i+1}]$, $i\in[1,length(X)-2]$. The weight of the 1D convolution is $W^{'}=[w_1,w_2,w_3]$,  and the convolution process is as follows:\par

\begin{equation}
Y_i^{'} = W^{'}X_i^{'T}=w_1X_{i-1}+w_1X_{i}+w_1X_{i+1}
\end{equation}

where$Y_i^{'}$ is the output after the convolution calculation and the number of channels is kept the same as $X_i$. From equations (1) and (2), it can be seen that the traditional 1D convolution calculation is equivalent to the 1D convolution calculation using time superposition. That is to say, by embedding the time superposition operation into the 2D convolution, the 2D convolution can process 3D data. At this time, the computation process and the number of parameters of the 2D convolution are then the same as those of the 3D convolution.\par

However, as can be seen from the above time stacking methods that stacking all channels of adjacent time data directly can establish effective time relationship, but it will cause a sharp increase in the amount of calculation and parameters. In order to balance efficiency and accuracy, we adopt the method of channel partial stacking to ensure that the number of channels remains constant before and after the superposition.\par

On the other hand, in the actual processing process, it is found that the spatial-temporal relationship of adjacent data could not be fully explored by using simple superposition method only, so we use the channel crossover method to mix the channel information of adjacent spatial-temporal data to establish a more effective spatial-temporal relationship. Channel crossover refers to the cross-mixing of channel data in a comb-like pattern in the channel dimension of adjacent temporal data, and we use partial channel crossover for data mixing in order to ensure the efficiency of the model.\par

Let $X_{i-1}=(x_1^{i-1},x_2^{i-1},...,x_n^{i-1}|n\in N)$, $X_{i}=(x_1^{i},x_2^{i},...,x_n^{i}|n\in N)$, $X_{i+1}=(x_1^{i+1},x_2^{i+1},...,x_n^{i+1}|n\in N)$, $i\in[1,length(X)-2]$. where $N$  is the number of channels, then the temporal superposition crossover process is:\par

\begin{equation}
\begin{aligned}
X_i^{'}&=(x_1^{i-1},x_2^{i},x_3^{i+1},...,x_{n-2}^{i-1},x_{n-1}^{i},x_n^{i+1}|n\in N)\\
&=X_{i-1}\copyright X_i\copyright X_{i+1}
\end{aligned}
\end{equation}

where $\copyright $ is the channel crossover operation and $X_i^{'}$ is the output after the temporal channel crossover operation.\par

\subsection {ResBlockT}

The TSCM proposed in this paper is a zero-parameter, zero-computation, plug-and-play module that can be combined with 2D convolution to make 2D convolution capable of processing 3D data. Taking the ResNet network as an example, we replace the normal convolution of the residual branch in ResBlcok with a hybrid convolution of "TSCM+2D convolution" to obtain a new block called ResBlockT, as shown in Figure 3.\par

Let the input feature of ResBlockT be $f^{T\times N\times H\times W}$, where $T$ is the timing length, $N$ is the number of channels and $H\times W$ is the resolution of the feature map. Taking the ResBlockT with bottleneck structure as an example, the channel information is first mixed by TSCM on adjacent data, and let its processing be $TSCM(\cdot)$. Then the channel compression is performed by a 2D convolution with a convolution kernel as follows:\par

\begin{equation}
f_1^{T\times \frac{N}{4}\times H\times W}=Covn_{1\times 1}(TSCM(f^{T\times N\times H\times W}))
\end{equation}

Where $f_1^{T\times \frac{N}{4}\times H\times W}$ is the output after channel compression. Repeating the above process to process $f_1^{T\times \frac{N}{4}\times H\times W}$ through the TSCM and then processing it using a 2D convolution with a convolution kernel of $3\times 3$:\par

\begin{equation}
f_2^{T\times \frac{N}{4}\times H\times W}=Covn_{3\times 3}(TSCM(f_1^{T\times \frac{N}{4}\times H\times W}))
\end{equation}

where $f_2^{T\times \frac{N}{4}\times H\times W}$ is the output data. Finally the feature is up-dimensioned and short-connected summed after passing the TSCM.\par

\begin{equation}
\begin{aligned}
f_{result}^{T\times N\times H\times W}&=Covn_{1\times 1}(TSCM(f_2^{T\times \frac{N}{4}\times H\times W}))\\
&+f^{T\times N\times H\times W}
\end{aligned}
\end{equation}

where $f_{result}^{T\times N\times H\times W}$ is the final output of the residual block. The above is the process of calculating ResBlockT for a bottleneck structure, and the process of calculating ResBlockT for a non-bottleneck structure is similar to this.\par

\begin{table*}[!htbp]
\centering
\caption{We compare the performance on the RWTH with different CSLR models, with WER as the metric (lower is better), where “Full” means that only the full RGB image is used for recognition, and “Extra clues” means that other cues are used for recognition}
\label{tab:aStrangeTable1}
\begin{tabular}{c|c|c|c|c|c}
\hline  
\multirow{2}{*}{Methods}& \multirow{2}{*}{Backbone}& \multirow{2}{*}{Full}& \multirow{2}{*}{Extra clues}& \multicolumn{2}{c}{WER(\%)}\\
\cline{5-6}
 & & & & Dev& Test\\
\hline
Re-Sign\cite{koller2017re}& GoogLeNet& Y& -& 27.1& 26.8\\
\hline  
SFL\cite{niu2020stochastic}& ResNet18& Y& -& 26.2& 26.8\\
\hline  
CNN+LSTM+HMM\cite{koller2019weakly}& GoogLeNet& -& Y& 26.0& 26.0\\
\hline  
FCN\cite{rastgoo2021sign}& Custom& Y& -& 23.7& 23.9\\
\hline  
DNF\cite{cui2019deep}& GoogLeNet& -& Y& 23.1& 22.9\\
\hline  
CMA\cite{pu2020boosting}& GoogLeNet& -& Y& 21.3& 21.9\\
\hline  
VAC\cite{min2021visual}& ResNet18& Y& -& 21.2& 22.3\\
\hline  
STMC\cite{zhou2020spatial}& VGG11& -& Y& 21.1& 20.7\\
\hline  
MSTNet\cite{zhu2022multi}& ResNet34& Y& -& 20.3& 21.4\\
\hline  
TLP\cite{hu2022temporal}& ResNet18& Y& -& 19.7& 20.8\\
\hline  
HST-GNN\cite{kan2022sign}& ResNet152& -& Y& 19.5& 19.8\\
\hline  
H-GAN\cite{elakkiya2021optimized}& Custom& -& Y& 18.8& 20.7\\
\hline  
ResNetT34& ResNet34& Y& -& 21.1& 21.1\\
\hline  
\end{tabular}
\end{table*}

This paper improves the ResNet34 network based on the bottleneck structure ResBlockT. The new network is called the ResNetT34 network and its network architecture is shown in Figure 1. ResNetT34 consists of two parts. The first part is the original ResNet34 network, including the large convolution kernel down-sampling of ResNet34 network, res2, res3 and part of res4. During training, the network layer of this part is trained using the method of random gradient descent. The second part consists of ResBlockT replacing the last seven ResBlocks of the original ResNet34 network, including the last four ResBlocks in res4 and the three ResBlocks in res5, with two down-sampling operations in the temporal dimension, and finally the whole network is trained using multi-level CTC loss.\par

\section{Experiment}

In this section, we conduct comprehensive experiments on two widely used CSLR datasets to validate the effectiveness of the proposed model in this paper. A series of ablation experiments are also carried out to demonstrate the role of each component of the proposed model.\par

\subsection {Dataset}

RWTH-PHOENIX-Weather-2014(RWTH) dataset\cite{koller2015continuous}: RWTH is recorded by a public weather radio and television station in Germany. All the presenters were dark clothes and performed sign language in front of a clean background. The videos in this dataset are recorded by 9 different presenters, and there are 6841 different sign language sentences in total (including 77321 sign language word instances and 1232 words). All videos are preprocessed to a resolution of $210\times 260$, and the frame rate is 25 frames per second (FP/S). The dataset is officially divided into 5672 training samples, 540 validation samples and 629 test samples.\par

Chinese Sign Language(CSL) dataset\cite{huang2018video}: CSL contains 100 Chinese daily expressions, each sentence is demonstrated 5 times by 50 presenters, and the vocabulary size is 178. The video resolution is $1280\times 720$, and the frame rate is 30 FP/S. The dataset includes two ways to divide training set and test set: split \RNum{1} and split \RNum{2}. This paper is an experimental study on split \RNum{2}, which divides 100 sentences into two parts, 94 sentences as the training set and the other 6 sentences as the test set.\par

\subsection {Implementation details}

In the overall model of this paper, Adam\cite{kingma2014adam} optimizer is used for training, with the initial learning rate and weight factor set to $10^{-4}$, and the batch size used is 2. When training with the RWTH dataset, random cropping and random flipping are used for data enhancement. For random clipping, the input data size is $256\times 256$, and the size after random clipping is $224\times 224$, to fit the input shape of the model. For random flipping, the flipping probability is set to 0.5. The flipping and cropping processes are carried out for the video sequences. In addition, a temporal enhancement process is performed to grow or shorten the length of the video sequences randomly within $\pm 20\%$. Finally, the model is trained with 3-level CTC loss. For model testing, only the center clipping is used for data enhancement and a bundle search algorithm is used for decoding in the final CTC decoding stage with a bundle width of 10. There were 85 epochs in the training stage, and the learning rate decreased by 80\% at the 45th and 65th epochs. When training with the CSL dataset, only random clipping is used for data enhancement and the model is trained using a 2-level CTC loss. For testing, only central cropping is used for data augmentation, with a total of 40 epochs in the training phase, and the learning rate is reduced by 90\% at the 30th epoch. The graphics card used in this experiment is RTX3090Ti with 24G of GPU-specific memory.\par

\subsection {Evaluation criteria}

Word Error Rate(WER)\cite{koller2015continuous}, as the evaluation standard, is widely used in CSLR. It is the sum of the minimum number of insertion operations, substitution operations, and deletion operations required to convert a recognized sequence into a standard reference sequence. Lower WER means better recognition performance, and its definition is as follows:\par

\begin{equation}
WER=100\%\times \frac{ins+del+sub}{sum}
\end{equation}

where $ins$ indicates the number of words to be inserted, $del$ indicates the number of words to be deleted, $sub$ indicates the number of words to be replaced, and $sum$ indicates the total number of words in the label.\par

\subsection {Experimental results}

The proposed method is experimented on two publicly available datasets, the RWTH and the CSL, respectively, and the experimental results are shown in Tables \RNum{1} and \RNum{2}. The curves generated from the WER in Table \RNum{1} and \RNum{2} are shown in Figure 4 and 5. As can be seen in Table \RNum{1}, the model proposed in this paper achieves highly competitive result on the RWTH compared to other state-of-the-art models, with a WER of 21.1\% on both the validation and test sets. Table \RNum{2} shows that the proposed model also achieves competitive result on the CSL, with a WER of 26.4\% on the test set. It can be seen in Figure 4 and Figure 5 that the WER decreases with the increase of epoch. When the first “lr” changes, the WER decreases significantly. On the RWTH, the WER reaches the minimum value of 21.1\% at the 66th epoch for the validation set, and 21.1\% at the 48th epoch for the test set; On the CSL, the WER of the test set reaches the minimum value of 26.4\% at the 37th epoch.\par

\begin{table*}[!htbp]
\centering
\caption{We partition the CSL dataset using the spilt II method and conducted experiments to compare the performance with different CSLR models, with WER as the metric (lower is better), where “Full” indicates that only the full RGB image is used for recognition and “Extra clues” indicates that other cues are used for recognition}
\label{tab:aStrangeTable2}
\begin{tabular}{c|c|c|c}
\hline  
Methods& Full& Extra clues& WER(\%)\\
\hline
HRNE\cite{pan2016hierarchical}& Y& -& 63.0\\
\hline
CTM\cite{guo2019connectionist}& Y& -& 61.9\\
\hline
HLSTM\cite{guo2018hierarchical}& Y& -& 48.7\\
\hline
DenseTCN\cite{guo2019dense}& Y& -& 44.7\\
\hline
Align-iOpt\cite{pu2019iterative}& Y& -& 32.7\\
\hline
STMC\cite{zhou2020spatial}& -& Y& 28.6\\
\hline
HST-GNN\cite{kan2022sign}& -& Y& 27.6\\
\hline
SBD-RL\cite{wei2020semantic}& Y& -& 26.8\\
\hline
CMA\cite{pu2020boosting}& -& Y& 24.5\\
\hline
ResNetT34& Y& -& 26.4\\
\hline  
\end{tabular}
\end{table*}

\begin{figure*}
  \begin{minipage}{0.5\textwidth}
\includegraphics[width=3.5in,height=2.33in]{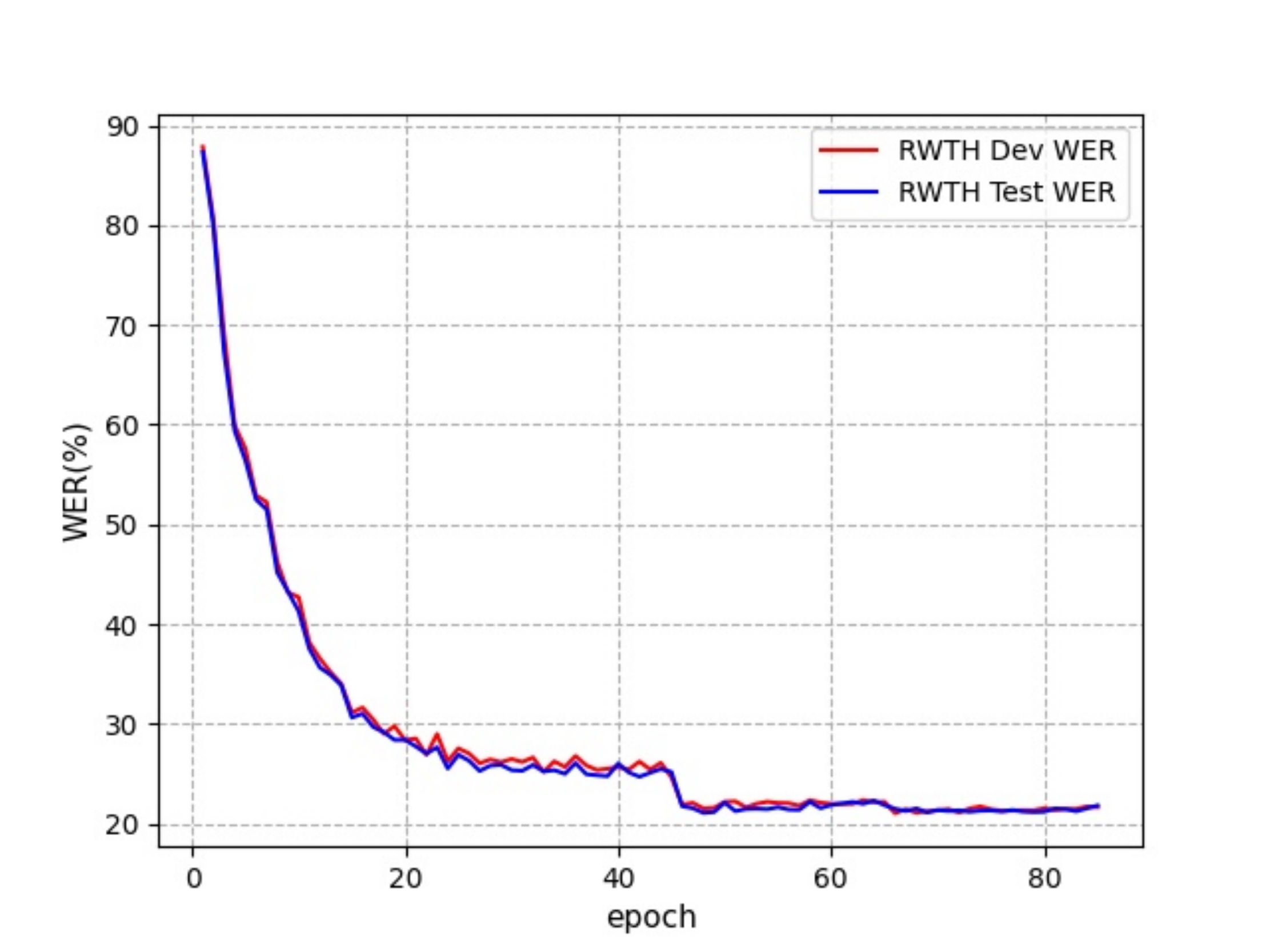}
\caption{WER variation curves of the RWTH validation and test sets.}
\label{fig:ljxy4}
\end{minipage}
\hfill
\begin{minipage}{0.5\textwidth}
\includegraphics[width=3.5in,height=2.33in]{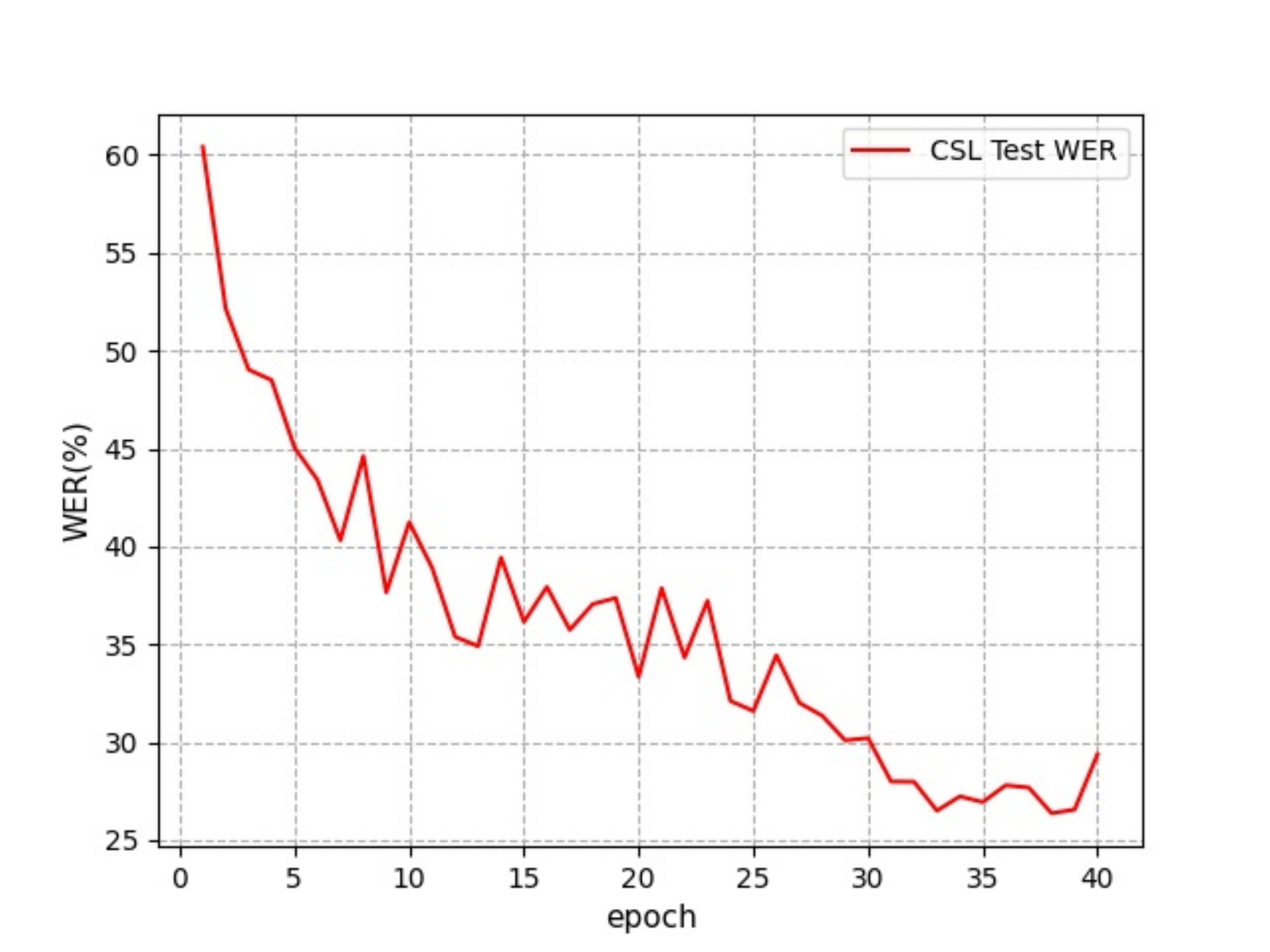}
\caption{WER variation curve of the CSL test set.}
\label{fig:ljxy5}
\end{minipage}
\end{figure*}

\subsection {Real-time and deploy-ability study}

To further analyze the superiority of the proposed model over other spatial-temporal convolutions in terms of real-time and deploy-ability, we will conduct comparative experiments on 2D convolution, TSCM+2D convolution, 2+1D convolution and 3D convolution in terms of the number of parameters, computational effort, inference time and accuracy.\par

In the comparison experiments we use ResNetT34 as the backbone and substitute the last 7 layers of the Block using different convolutions. The computation and inference time of the models are strongly correlated with the size of the input data. We set the input data size to $3\times 224\times 224$ and the timing length to 200 frames, calculate the number of parameters and computation for the different models under the above conditions. For each model, we count the inference time for 20 consecutive times and calculate its average value as the criterion, and carry out experimental validation on the RWTH, with WER as the accuracy indicator, where a smaller WER means higher accuracy. The experimental results are shown in Table \RNum{3}, while the data from Table \RNum{3} are visualized in Figure 6 and Figure 7.\par

It can be seen from Table \RNum{3} that: 1) The 2D convolution and TSCM+2D convolution are consistent in terms of parameter amount and calculation amount, but the inference time of 2D convolution is 8.5ms higher than that of TSCM+2D convolution, which is due to the fact that TSCM is all shift operations, which only increases the delay time without increasing the number of parameters and computation; 2) The accuracy of the model obtained by using 3D convolution is improved by 1.4\% compared to that obtained by using TSCM+2D convolution, but in terms of the number of parameters and computation, 3D convolution is 160.9\% and 74.3\% more than TSCM+2D convolution respectively, and the inference time is increased by 19.5ms; 3) Compared with TSCM+2D convolution, 2+1D convolution only increases the inference time of the model by 1.8ms, but its parameters increase by 28.6\%, the calculation amount increases by 12.7\%, and the model accuracy decreases by 1.4\%.\par

In summary, using our proposed "TSCM+2D convolution" hybrid convolution results in a smaller loss in model accuracy compared to 3D convolution, while the number of parameters, computation and inference time are better than other spatial-temporal convolutions, achieving a balance between performance and accuracy. As can be seen in Figure 6, the "TSCM+2D convolution" hybrid convolution has a smaller number of parameters and computational effort, which makes the model less expensive to deploy, and the inference time is reduced compared to other spatial-temporal convolutions, thus ensuring the real-time performance of the model.\par

\begin{table*}[!htbp]
\centering
\caption{Comparison of the number of parameters, computation cost, inference time and WER on the RWTH test set for different convolutions}
\label{tab:aStrangeTable3}
\begin{tabular}{c|c|c|c|c}
\hline  
Type of convolution& 2D& TSCM+2D& 2+1D& 3D\\
\hline
Number of parameters(M)& 22.0& 22.0& 28.3& 57.4\\
\hline
Parameters memory(MB)& 83.9& 83.9& 108.0& 219.0\\
\hline
Computational cost(GFlops)& 671.1& 671.1& 756.3& 1170.0\\
\hline
Inference time(ms)& 73.4& 81.9& 83.7& 101.4\\
\hline
WER(\%)& 34.0& 21.1& 21.4& 20.8\\
\hline  
\end{tabular}
\end{table*}

\begin{figure*}
  \begin{minipage}{0.5\textwidth}
\includegraphics[width=3.5in,height=2.33in]{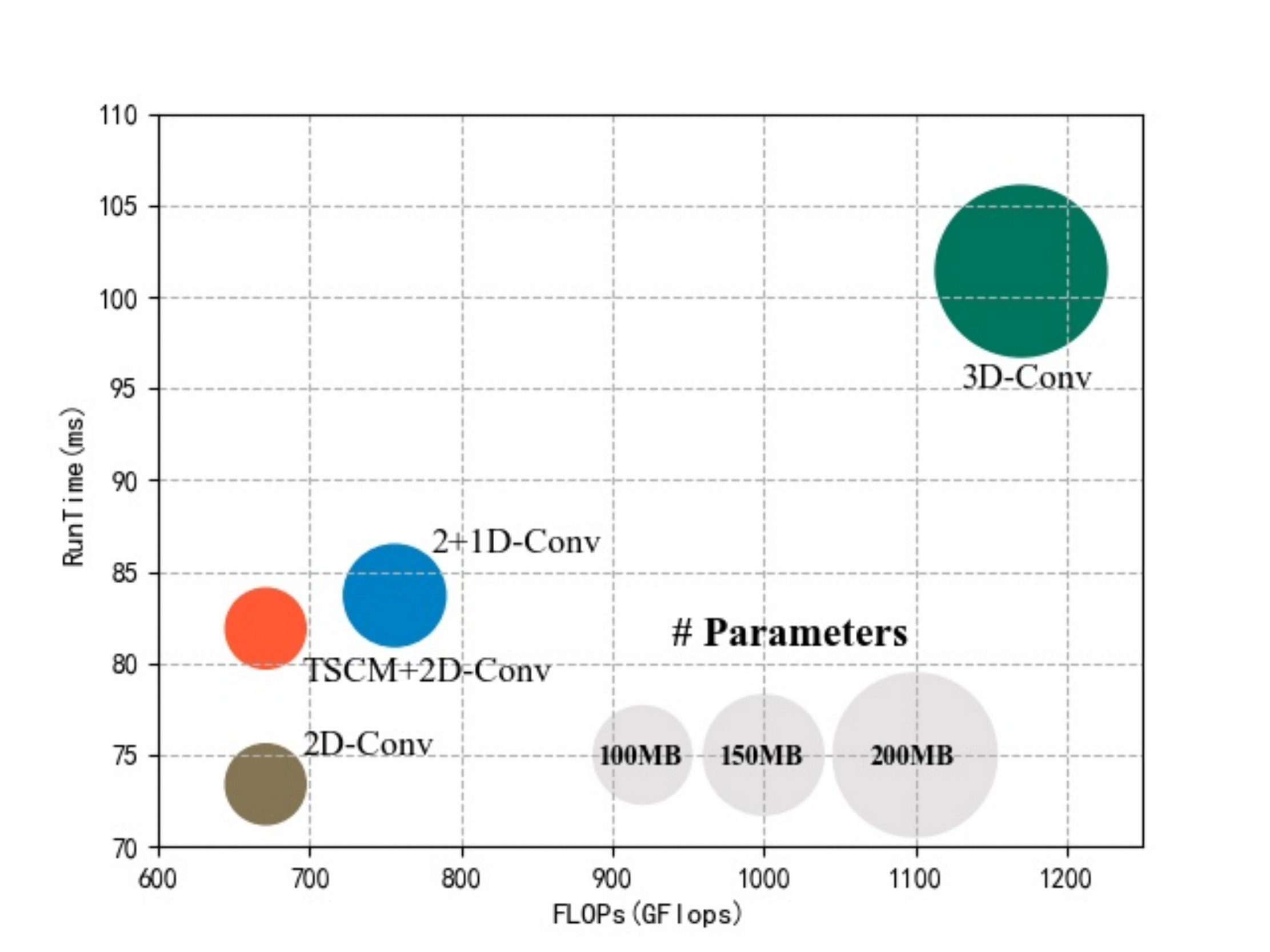}
\caption{Scatter plots of different convolutional models in terms of number of parameters, computational cost and inference time.}
\label{fig:ljxy6}
\end{minipage}
\hfill
\begin{minipage}{0.5\textwidth}
\includegraphics[width=3.5in,height=2.33in]{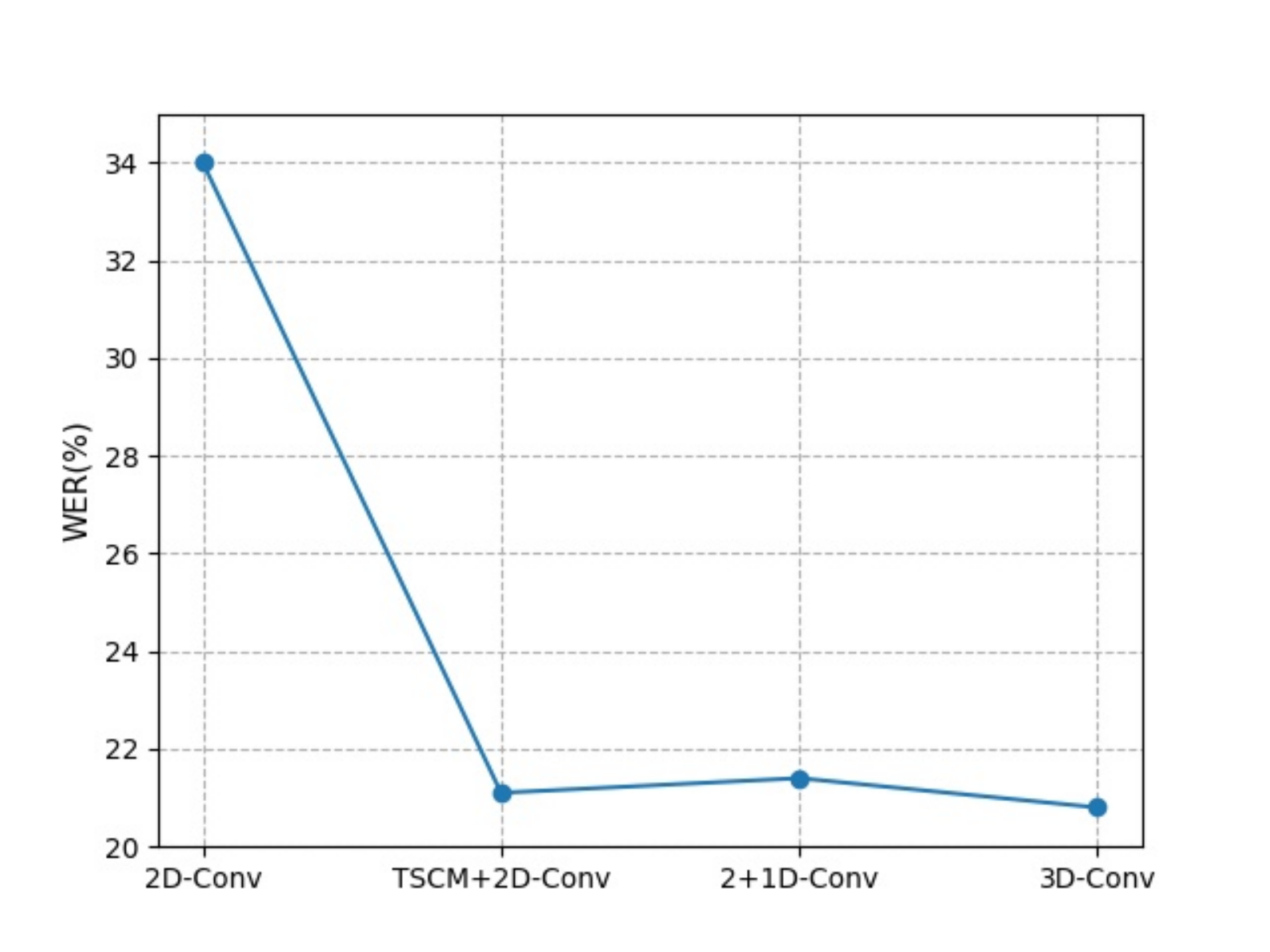}
\caption{ Results of experimental validation of different convolutional models on the RWTH dataset (WER on the test set).}
\label{fig:ljxy7}
\end{minipage}
\end{figure*}

\begin{table*}[!htbp]
\centering
\caption{Effect of different ResBlockT layers on model accuracy}
\label{tab:aStrangeTable4}
\begin{tabular}{c|c|c}
\hline  
\multirow{2}{*}{Number of ResBlockT}& \multicolumn{2}{c}{WER(\%)}\\
\cline{2-3}
 & Dev& Test\\
\hline  
4& 21.7& 21.6\\
\hline
5& 21.3& 21.5\\
\hline
6& 21.4& 21.4\\
\hline
7& 21.1& 21.1\\
\hline
8& 21.0& 21.4\\
\hline  
\end{tabular}
\end{table*}

\begin{table*}[!htbp]
\centering
\caption{Effect of different ResNetT model size on model accuracy}
\label{tab:aStrangeTable5}
\begin{tabular}{c|c|c|c}
\hline  
\multirow{2}{*}{Model size}& \multicolumn{2}{c|}{RWTH}& \multirow{2}{*}{CSL}\\
\cline{2-3}
 & Dev(\%)& Test(\%)& \\
\hline  
ResNetT34& 21.1& 21.1& 26.4\\
\hline
ResNetT50& 20.1& 20.3& 44.0\\
\hline
ResNetT101& 20.3& 20.5& -\\
\hline  
\end{tabular}
\end{table*}

\subsection {Ablation experiment}
In this section, we conduct ablation experiments on the RWTH dataset using ResNetT34 as the baakbone to further validate the effectiveness of the individual components of the model, using WER as the metric in the ablation experiments, with a smaller WER representing better performance.\par

\textbf{Ablation of ResBlockT numbers}. In this paper, the proposed TSCM and 2D convolution are combined to form a "TSCM+2D convolution" hybrid convolution, which is used to replace the ordinary convolution of residual branches in ResBlock to obtain a new block, called ResBlockT. ResBlockT enables the ResNet model to have spatial-temporal modeling capabilities. Specifically, we use ResBlockT to replace ResBlocks in the ResNet model. The replacement principle is from the back to the front. As the number of replacement increases, the spatial-temporal modeling capabilities of the model are also different, as shown in Table \RNum{4}.\par

It can be seen from Table \RNum{4} that the spatial-temporal modeling capability of the model initially increases with the increase of the number of ResBlockT. When the number of ResBlockT reaches 7, the spatial-temporal modeling capability of the model reaches the maximum, at which point the WER on the test set is minimised, falling to 21.1\%. Later, when the number of ResBlockT continues to increase, the spatial-temporal modeling capability of the model decreases.\par

\begin{table*}[!htbp]
\centering
\caption{Effect of different superimposition and fusion methods on model accuracy}
\label{tab:aStrangeTable6}
\begin{tabular}{c|c|c}
\hline  
\multirow{2}{*}{Superimposition methods}& \multicolumn{2}{c}{WER(\%)}\\
\cline{2-3}
 & Dev& Test\\
\hline  
TSM& 21.6& 21.7\\
\hline
Time superimposition& 21.5& 21.2\\
\hline
Time superposition crossover& 21.1& 21.1\\
\hline
Time random superposition crossover& 28.0& 28.5\\
\hline  
\end{tabular}
\end{table*}

\begin{table*}[!htbp]
\centering
\caption{Effect of different partial channel time superimposed crossover modules on model }
\label{tab:aStrangeTable7}
\begin{tabular}{c|c|c|c}
\hline  
\multirow{2}{*}{Temporal superimposed crossover approaches}& \multicolumn{2}{c|}{WER(\%)}& \multirow{2}{*}{inference time (ms)}\\
\cline{2-3}
 & Dev& Test& \\
\hline  
1/3-channel temporal superimposed crossover& 21.1& 21.1& 81.9\\
\hline
1/5-channel temporal superimposed crossover& 21.0& 20.3& 85.2\\
\hline
1/7-channel temporal superimposed crossover& 21.9& 22.1& 88.9\\
\hline  
\end{tabular}
\end{table*}

\begin{table*}[!htbp]
\centering
\caption{Effect of the different number of time down-sampling on model accuracyl accuracy}
\label{tab:aStrangeTable8}
\begin{tabular}{c|c|c}
\hline  
\multirow{2}{*}{Number of time down-sampling}& \multicolumn{2}{c}{WER(\%)}\\
\cline{2-3}
 & Dev& Test\\
\hline  
0& 23.5& 23.4\\
\hline
1& 23.2& 23.5\\
\hline
2& 21.1& 21.1\\
\hline
3& 25.8& 25.6\\
\hline  
\end{tabular}
\end{table*}

\begin{table*}[!htbp]
\centering
\caption{Effect of the different number of CTC loss level on model accuracy}
\label{tab:aStrangeTable9}
\begin{tabular}{c|c|c}
\hline  
\multirow{2}{*}{Number of CTC loss level}& \multicolumn{2}{c}{WER(\%)}\\
\cline{2-3}
 & Dev& Test\\
\hline  
1-level CTC loss& 24.2& 23.1\\
\hline
2-level CTC loss& 22.3& 22.1\\
\hline
3-level CTC loss& 21.1& 21.1\\
\hline  
\end{tabular}
\end{table*}

\textbf{Ablation of ResNetT model size}. The ablation effect of model size is shown in Table \RNum{5}, where changing the size of the network model layers while keeping the number of ResBlockT constant essentially changes the spatial feature extraction capability of the model. On RWTH and CSL datasets, the performance of the same model on the two datasets may be different. On the CSL dataset, the WER on the test set is 26.4\% when the model is ResNetT34, and 44.0\% when the model is ResNetT50, a 17.6\% reduction in WER. In contrast, on the RWTH dataset, ResNetT50 give a 0.8\% improvement in WER over ResNetT34. Thus, for the CSL dataset, it is not necessary to extract the spatial features of an oversized model in the early stage, while for the RWTH dataset, better spatial feature extraction capability can effectively improve the performance of the model. On the RWTH dataset, when the model is ResNetT50, it achieves a WER of 20.3\% on the test set, which is a state-of-the-art result for a model using only RGB images as input, and a 0.5\% improvement over TLP\cite{hu2022temporal}. On the CSL dataset, when the model is ResNetT34, the WER of its test set reaches 26.4\%. In the model that only uses RGB images as input, the experimental results reach the most advanced level, 0.4\% higher than SBD-RL\cite{wei2020semantic}.\par

\textbf{Ablation of different superimposition methods}. The ablation effects of the different superimposition methods are shown in Table \RNum{6}. We replace only the TSCM in the model, keeping the other components unchanged. In the ablation study, it is found that the TSM\cite{lin2019tsm} method increased the WER by 0.5\% on the validation set and by 0.6\% on the test set relative to the temporal overlay method. A simple overlay of the temporal data in the channel dimension also performs well, but increases the WER by 0.4\% on the validation set and by 0.1\% on the test set relative to the overlay crossover approach. We also validate the operation of random channel crossover on the time-series data and find that random channel crossover can reduce the spatial-temporal modelling capability of the model, increasing the WER by 6.9\% on the validation set and 7.4\% on the test set.\par

\textbf{Ablation of partial channel time superimposed crossover modules}. The ablation effect of the partial channel time overlay crossover module is shown in Table \RNum{7}. The number of channels remains constant before and after the temporal data overlay crossover operation, so the parameter partial channels expresses not only the percentage of individual data channels, but also the number of adjacent temporal data taken. The 1/7 channel temporal superimposed crossover results in a 0.8\% increase in WER on the validation set and a 1.0\% increase in WER on the test set compared to the 1/3 channel temporal superimposed crossover. The difference in performance between 1/5-channel temporal superimposed crossover and 1/3-channel temporal superimposed crossover is not significant, but the inference time of 1/5-channel temporal superimposed crossover is 3.3ms longer than that of 1/3-channel temporal superimposed crossover. So on balance, the 1/3-channel time superimposed crossover is an optimal choice.\par

\textbf{Ablation of the number of time drop samples}. The ablation effect of the number of time down-sampling is shown in Table \RNum{8}. The network uses 2 time down-sampling as shown in Figure 1, and the number of downs-ampling is calculated from back to front, using 1 time down-sampling, retaining the last 1 layer of 1D-MaxPool; using 2 time down-sampling, retaining 2 layers of 1D-MaxPool; using 3 time down-sampling, on the basis of 2 time down-sampling, and then adding 1 layer of 1D-MaxPool between model Part1 and Part2. It can be seen from Table \RNum{8} that with the increase of the number of down-samples, the WER obtained decreases first and then increases. When the number of down-samples is 2, the performance of the model is optimal, and the WER obtained on the test set reaches 21.1\%.\par

\textbf{Ablation of multi-level CTC loss}. The ablation effect of multi-level CTC loss is shown in Table \RNum{9}. In this paper, we use 3-level CTC loss, and its location is shown in Figure 1. It can be seen from Table \RNum{9} that with the increase of CTC loss level, WER is in a downward trend. When the CTC loss level is 3, the performance of the model reaches the optimal level, and the WER of the test set reaches 21.1\%.\par

\section{Conclusion}
In recent years, research in CSLR has focused on reducing the WER. The complexity of the model can make the WER lower and lower on the one hand, but on the other hand the number of model parameters and the amount of computation it brings is increasingly large. The ultimate goal of CSLR is to solve the communication barrier between special and normal people, so the real-time and deploy-ability of the model is required. In this paper, we propose a new zero-parameter, zero-computation temporal superposition crossover module, which is combined with 2D convolution to form a "TSCM+2D convolution" hybrid convolution, which can well establish spatial-temporal dependencies when performing video recognition. The hybrid convolution is applied to the ResBlock of the ResNet network to form the new ResBlockT. The improved ResNetT network has good spatial-temporal modelling capability. Experiments show that the CSLR model established in this paper outperforms other spatial-temporal convolutions in terms of the number of parameters, computation and inference time, although it has a smaller loss in accuracy compared to the 3D convolution-based model. The reduction in the number of parameters and computation reduces the deployment cost of the model, and the reduction in inference time ensures the real-time performance of the model, achieving a balance between performance and accuracy. In addition, the model built in this paper is the first end-to-end network model with pure 2D convolution in CSLR.\par

CSLR aims to address the communication problem between hearing impaired people and normal people, and the computation and parameters of the model need to meet the requirements of real-time and deploy-ability. The method proposed in this paper has improved the real-time and deploy-ability of the model to a certain extent, but in order to further improve the real-time and deploy-ability of the model, a feasible direction is to sparse the input data. How to use less and lower resolution input data to achieve the same or better recognition effect as the current is a problem worth studying.\par

\section*{Acknowledgment}

This work was supported in part by the Development Project of Ship Situational Intelligent Awareness System, China under Grant MC-201920-X01, in part by the National Natural Science Foundation of China under Grant 61673129. \par

~\\\par
\textbf{Data availability} The datasets used in the paper are cited properly.\par

\section*{Declarations}

\textbf{Conflict of interest} The authors declare that they have no known competing financial interests or personal relationships that could have appeared to influence the work reported in this paper.\par


%





\ifCLASSOPTIONcaptionsoff
  \newpage
\fi





\bibliographystyle{IEEEtran}
\bibliography{IEEEabrv,Bibliography}

\vfill


\end{document}